\documentclass[conference]{IEEEtran}
\IEEEoverridecommandlockouts
\usepackage{cite}
\usepackage{soul}
\usepackage{float}
\usepackage{amsmath,amssymb,amsfonts}
\usepackage{graphicx}
\usepackage{textcomp}
\usepackage{caption}
\usepackage{subcaption}
\usepackage{algorithm}
\usepackage{algpseudocode}
\usepackage{multirow}
\usepackage[table,xcdraw]{xcolor}
\usepackage{booktabs}
\usepackage[most]{tcolorbox}
\def\BibTeX{{\rm B\kern-.05em{\sc i\kern-.025em b}\kern-.08em
    T\kern-.1667em\lower.7ex\hbox{E}\kern-.125emX}}

\newcommand{\acronym}{\textit{SelfAct}}
    
\begin{document}

\title{\acronym: Personalized Activity Recognition based on Self-Supervised and Active Learning}

\author{\IEEEauthorblockN{Luca Arrotta, Gabriele Civitarese, Samuele Valente, Claudio Bettini}
\IEEEauthorblockA{\textit{EveryWare Lab, Dept. of Computer Science}\\\textit{University of Milan}\\ \{luca.arrotta, gabriele.civitarese, claudio.bettini\}@unimi.it} samuele.valente@studenti.unimi.it
}

\maketitle

\begin{abstract}
Supervised Deep Learning (DL) models are currently the leading approach for sensor-based Human Activity Recognition (HAR) on wearable and mobile devices. However, training them requires large amounts of labeled data whose collection is often time-consuming, expensive, and error-prone. 
At the same time, 
due to the intra- and inter-variability of activity execution,
activity models should be personalized for each user. 
In this work, we propose \acronym: a novel framework for HAR combining self-supervised and active learning to mitigate these problems. \acronym\ leverages a large pool of unlabeled data collected from many users to pre-train through self-supervision a DL model, with the goal of learning a meaningful and efficient latent representation of sensor data.  The resulting pre-trained model can be locally used by new users, which will fine-tune it thanks to a novel unsupervised active learning strategy. 
Our experiments on two publicly available HAR datasets demonstrate that \acronym\ achieves results that are close to or even better than the ones of fully supervised approaches with a small number of active learning queries.
\end{abstract}


\section{Introduction}
The recent research on sensor-based Human Activity Recognition (HAR) on mobile/wearable devices is dominated by solutions based on Deep Learning (DL) approaches~\cite{wang2019deep}.
Supervised learning is the most commonly proposed approach in this area since is capable of reaching significantly high recognition rates~\cite{ronald2021isplinception}.

However, such approaches require a huge amount of labeled data to create models capable of generalizing. Indeed, the high intra- and inter-variability of activity execution is a significant challenge in this domain, and personalized models (i.e., models trained only with data from the target users) are the ones performing the best~\cite{weiss2012impact}.  

However, annotation is costly, time-consuming, intrusive, and thus often prohibitive. For this reason, several research groups are investigating solutions for HAR requiring limited labeled datasets~\cite{presotto2022semi}.

Among these approaches, self-supervised learning methods are emerging~\cite{khaertdinov2021contrastive,jain2022collossl,haresamudram2022assessing}. The goal of self-supervised learning is to leverage large corpora of unlabeled data to learn to extract effective features from sensor data. 
While self-supervised approaches showed their effectiveness on NLP and computer vision tasks~\cite{baevski2022data2vec}, their application to sensor-based HAR in real-world scenarios is still an open research question.

One of the major drawbacks of existing approaches is that they assume that a small labeled dataset for fine-tuning is available. However, those works do not investigate how this dataset can be obtained in practice.

In the literature, active learning has been widely investigated to obtain annotated data samples in HAR data scarcity scenarios~\cite{hossain2019active,presotto2022semi, abdallah2012streamar}. However, those approaches assume that a pre-training labeled dataset is available to initialize the recognition model. Indeed, a query is triggered based on the confidence level of the classifier.
Applying the concept of active learning to a self-supervised setting is not trivial, since it requires deciding when to trigger queries by analyzing an unlabeled data stream.

In this work, we propose \acronym: a novel framework bridging the gap between self-supervised learning and active learning. \acronym\ relies on self-supervised learning to learn, from a large number of users, an effective feature representation of sensor data. Then, each user who wants to use \acronym\ downloads this pre-trained model on personal mobile/wearable devices. After the first phase where the personal device accumulates unlabeled sensor data, a clustering algorithm on the embeddings generated by the pre-trained model is performed. After the accumulation phase, \acronym\ analyzes in real-time each data sample in the stream and matches it with the closest cluster. A novel unsupervised active learning strategy, based on cluster density, is adopted to decide whether to trigger a query or not. Labeled data samples obtained through active learning are finally used to fine-tune the pre-trained self-supervised model and personalize it on the specific user.

The contributions of this work are the following:
\begin{itemize}
    \item We propose \acronym: a framework for personalized sensor-based HAR on mobile/wearable devices based on self-supervised learning.
    \item We design and integrate into \acronym\ a novel unsupervised active learning strategy to obtain labeled data samples. These samples are used to fine-tune a self-supervised model. 
    \item Our experiments on two public datasets show that \acronym\ achieves similar or even better recognition rates than fully supervised approaches with a very limited amount of active learning queries.
\end{itemize}


\section{Related work}
Most of the existing works in the literature tackling sensor-based HAR on mobile/wearable devices are based on supervised Deep Learning (DL) models~\cite{wang2019deep}. However, during their learning process, such models require large amounts of labeled data that are often prohibitive to acquire~\cite{chen2021deep}. Moreover, activity models should be personalized for each user to capture her peculiarities in activity execution~\cite{chen2021deep}. Several research groups are indeed working hard to mitigate the data scarcity problem, by proposing several categories of solutions.

Some research efforts proposed solutions based on automatic data annotation. Some of these works assume the existence of previously annotated datasets to pre-train an activity model, that is used for annotation~\cite{hassan2021autoact}. Other approaches require auxiliary data sources in the environment (e.g., acoustic data from microphones) to automatically annotate activity data~\cite{chatterjee2020laso}.
Finally, other works proposed knowledge-based heuristic approaches to generate weak labels (e.g., combining step count and GPS data)~\cite{cruciani2018automatic}. However, the scalability of such an approach is questionable since the heuristic was evaluated considering data collected by monitoring a single user. Moreover, the proposed method can not be used for data annotation when multiple activities share similar patterns in terms of step count and GPS data (e.g., standing and sitting).

Some research groups focused on solutions based on transfer learning \cite{sanabria2021unsupervised, soleimani2021cross}. However, such methods rely on models that are pre-trained with significant amounts of labeled data available in a source domain and then fine-tuned in the target domain with a few annotated samples.

Another category of solutions to mitigate the data scarcity problem is semi-supervised learning~\cite{abdallah2018activity}. However, these approaches still assume the existence of a small labeled dataset to initialize an activity classifier, that is then incrementally updated with samples of the unlabeled data stream annotated through techniques like self-learning, co-learning, active learning \cite{adaimi2019leveraging, hossain2019active, cui2022reinforcement}, and label propagation \cite{presotto2022semi}. 

Unsupervised approaches have also been proposed to tackle data scarcity. For instance, some research groups directly relied on clustering techniques to distinguish the different user's activities \cite{trabelsi2013unsupervised, kwon2014unsupervised, abedin2020towards, ahmed2022clustering}. However, they did not consider how to match clusters with the actual activity classes the user performs in realistic scenarios. 

More recently, self-supervised learning has been studied to efficiently learn (without supervision) meaningful features from sensor data for both mobile~\cite{saeed2019multi, tang2020exploring, almaslukh2017effective} and smart home~\cite{bouchabou2021using} activity recognition. The majority of existing solutions  assume the availability of a labeled dataset to fine-tune the self-supervised model and obtain the final activity classifier. 
A closely related work~\cite{ma2021unsupervised} uses a labeled validation dataset to match activities with clusters identified by k-means on self-supervised embeddings. Different from this work, \acronym\ does not assume the knowledge of the number of activities. Also, \acronym\ includes an unsupervised active learning strategy to select the samples representing each cluster to label only a few data samples through active learning. 
Overall, the major drawback of these works based on self-supervised learning is that they do not clarify how fine-tuning datasets can be obtained in realistic settings.

Finally, the work in~\cite{hiremath2022bootstrapping} considered BERT to extract from smart home sensor data a latent representation of the action units performed by the resident (e.g., a movement in the kitchen) that are then clustered. 
Hence, frequently occurring patterns of action units are found through a motif discovery process. The most frequent motifs are finally labeled by the resident thanks to active learning.
However, NLP-based techniques like BERT and motif discovery work well for discrete environmental sensors, but they cannot be directly applied to the raw inertial measurements collected by mobile and wearable devices.

\section{The \acronym\ framework}
\begin{figure*}[ht!]
    \centering
    \includegraphics[width=0.9\linewidth]{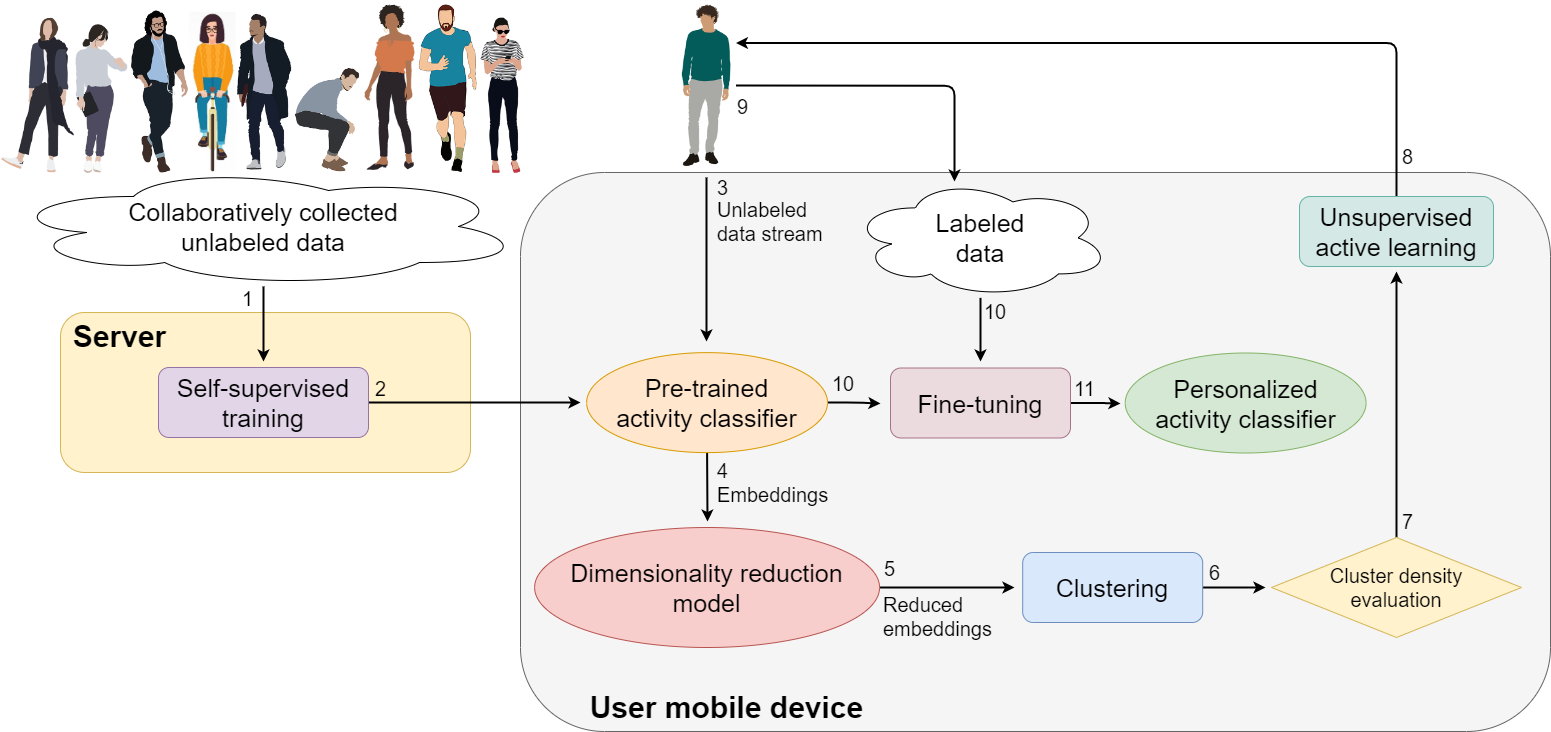}
    \caption{The overall architecture of \acronym}
    \label{fig:architecture}
\end{figure*}

In this section, we present \acronym:  our framework that generates a personalized activity classifier by combining self-supervised learning and unsupervised active learning. 

As depicted in Figure \ref{fig:architecture}, in the first phase (called \textit{self-supervised phase}), \acronym\ requires huge amounts of unlabeled inertial sensor data automatically and collaboratively collected from the mobile/wearable devices of multiple users (e.g., volunteers). This unlabeled dataset is used to train, in a self-supervised fashion, a deep neural network capable of extracting effective latent representations (i.e., embeddings) from sensor data. This process is described in detail in Section~\ref{sec:self_supervised_phase}.

In a second phase (called \textit{accumulation phase}), new users start using \acronym\ by downloading and deploying the pre-trained self-supervised model on their devices. 
Each user locally accumulates unlabeled inertial sensor data for a certain period. At the end of the \textit{accumulation phase}, such unlabeled data are transformed into embeddings with the pre-trained model and then used to train a personal dimensionality reduction model. Finally, 
the reduced embeddings are clustered. Intuitively, each cluster would likely represent a personal sensor data pattern. More details about this \textit{accumulation phase} are presented in Section~\ref{sec:accumulation_phase}.


After the \textit{accumulation phase}, \acronym\ starts the \textit{active learning phase} (presented in detail in Section~\ref{sec:active_learning_phase}).
This phase analyzes, in real-time, each new sensor data sample in the stream to decide whether to ask the user for feedback about the corresponding label (i.e., the activity that the user is actually performing). Intuitively, \acronym\ aims at asking for a label only 
whenever a sample is recognized as 
representative of a cluster.


Thanks to active learning, \acronym\ automatically collects a small amount of labeled data 
specific to the subject
and this data is used during the \textit{fine-tuning phase} (presented in Section \ref{sec:fine_tuning_phase}) to obtain a personalized activity classifier.     

The \acronym's algorithm running locally on clients is summarized in Algorithm~\ref{alg:selfact}.
In the following, we will describe each phase of \acronym\ in detail.

\begin{algorithm}[]
\footnotesize
\caption{Client algorithm of \acronym}
\label{alg:selfact}
\hspace*{\algorithmicindent} \textbf{Input:} the \textit{Feature Extraction Layers} $fe$ of the pre-trained self-supervised model, the accumulation threshold $ACC\_TH$  \\
\begin{algorithmic}[1]
\State $fth \gets$ randomly initialized layers of the \textit{Fine-Tuning Head}
\State $dr \gets$ NIL
\State $samplesNumber \gets 0$ 
\State $storage \gets \emptyset$
\State $reducedStorage \gets \emptyset$
\State $labeledSamples \gets \emptyset$
\State $clusters \gets \emptyset$
\For{$s$ data sample in the stream}
    \State $samplesNumber \gets samplesNumber+1$
    \State $emb_s \gets$ obtain embeddings using $fe$
    \State $storage \gets storage \bigcup \{emb_s\}$
    \If{$samplesNumber == ACC\_{TH}$}
        \State $dr \gets $ data reduction model trained using $storage$
        \State $reducedStorage \gets$ embeddings in $storage$ reduced using $dr$ 
        \State $clusters \gets $ apply clustering algorithm on $reducedStorage$
    \Else 
    \If{$samplesNumber > ACC\_{TH}$}
        \State $emb_s' \gets $ dimensionally reduced $emb_s$ using $dr$ model
        \State $matchingClust \gets $ closest cluster to $emb_s'$ in $clusters$
        \If{$ActiveLearningNeeded(emb_s', matchingClust)$}
            \State $l \gets$ feedback from user on the activity associated with $s$
            \State $labeledSamples \gets labeledSamples \bigcup \{(s,l)\}$
        \EndIf
        \State $matchingClust \gets matchingClust \bigcup \{emb_s'\}$
        \If{Fine-tuning conditions are verified}
            \State adjust $fth$ to cover all the classes in $labeledSamples$
            \State fine-tune $fe+fth$ using $labeledSamples$
            \State $fe+fth$ can be used to classify unlabeled samples
        \EndIf
    \EndIf
    \EndIf
\EndFor
\end{algorithmic}
\end{algorithm}

\subsection{Self-supervised phase}
\label{sec:self_supervised_phase}
The goal of the \textit{self-supervised phase} is to train, in a self-supervised fashion, a deep learning model capable of transforming raw sensor data samples into an efficient latent representation (i.e., embeddings). Intuitively, data samples representing the same activity pattern should be close in the latent space, while data samples from different activity patterns should be distant.

In this phase, \acronym\ relies on unlabeled raw sensor data that can be collaboratively and automatically collected from the mobile/wearable devices of several users. For the sake of this work, we do not focus on how such data are actually collected. For instance, such users can be volunteers involved in a large data collection campaign. Alternatively, unlabeled data could be collected by combining existing publicly available datasets for HAR.

\acronym\ pre-processes and segments the available unlabeled data into fixed-length segmentation windows that are used to train the self-supervised model. During the \textit{self-supervised phase}, this model is composed of two parts: i) \textit{Feature Extraction Layers} that learns how to extract embeddings from sensor data, and  ii) a \textit{Self-Supervised Head}, necessary to perform the self-supervised training process.
\acronym\ does not impose a specific self-supervised approach. As we will discuss in Section~\ref{sec:experiments}, in our experiments we implemented \acronym\ with a version of the SimCLR~\cite{chen2020simple} approach specifically adapted for the HAR domain~\cite{tang2020exploring}.


The output of the \textit{self-supervised phase} is a pre-trained deep-learning model only composed of the \textit{Feature Extraction Layers}, that is able to extract embeddings from unlabeled sensor data. This model can be locally deployed on the mobile devices of the new users that start using \acronym.

\subsection{Accumulation phase}
\label{sec:accumulation_phase}
The \textit{accumulation phase} starts when a new user downloads and locally deploys on her device the model pre-trained in the \textit{self-supervised phase} (i.e., \textit{Feature Extraction Layers}). The goal of the \textit{accumulation phase} is to accumulate from the user sufficient amounts of unlabeled data to reliably discover personal activity patterns through clustering.

\begin{figure}
    \centering
    \includegraphics[width=0.9\linewidth]{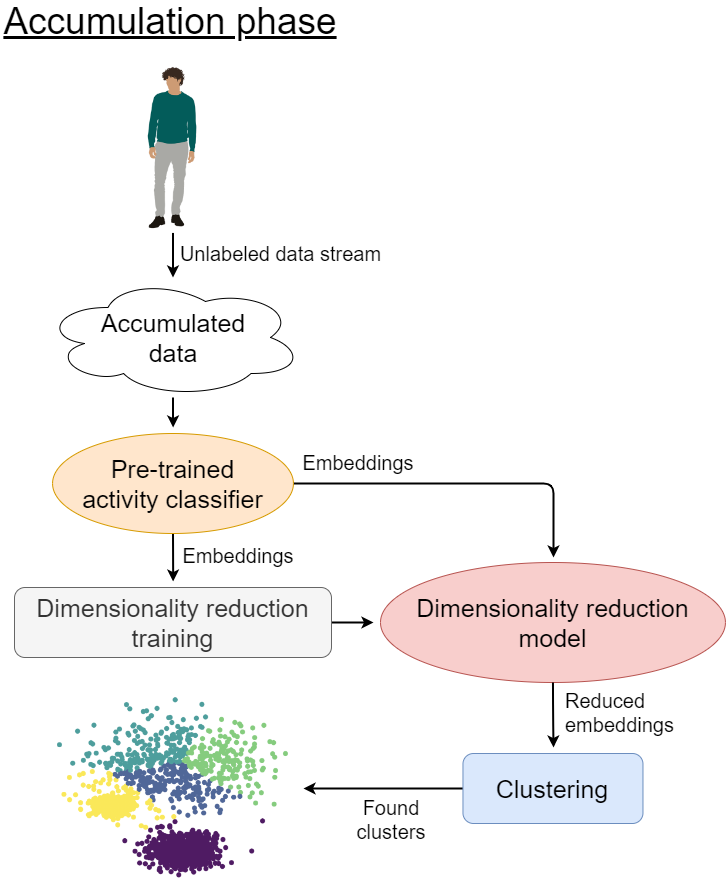}
    \caption{The \textit{accumulation phase} of \acronym}
    \label{fig:accumulation_phase}
\end{figure}

As depicted in Figure \ref{fig:accumulation_phase},
the user's mobile device starts collecting the stream of unlabeled inertial sensor data that is segmented into fixed-length windows until an arbitrary number of $ACC\_TH$ (accumulation threshold) samples is acquired. This parameter should be tuned based on the specific application, and, in real-world scenarios, its value should be high in order to collect unlabeled data over a sufficiently long time period (e.g., a day, or a week).

Once unlabeled data accumulation is completed, each sample is provided to the pre-trained self-supervised model to obtain embeddings. Then, to efficiently perform clustering to find activity patterns, we leverage all the generated embeddings to train a personalized dimensionality reduction model. Indeed, recent works demonstrated the impact of dimensionality reduction in improving the performance of clustering techniques~\cite{allaoui2020considerably}. Finally, the dimensionally reduced accumulated embeddings are provided to a clustering algorithm that does not make any prior assumption on the number of clusters (e.g., DBSCAN). The goal is to group in the same cluster those embeddings that have been generated while the user was performing the same activity pattern. We expect our method to obtain more than one cluster for each activity, since users may perform the same activity in different ways (e.g., walking indoors and walking outdoors).

\subsection{Active learning phase}
\label{sec:active_learning_phase}
The goal of the \textit{active learning phase} is to select, in real-time, data samples that should be labeled by the user. These labeled samples are then used by \acronym\ to fine-tune a personalized activity classifier.

\begin{figure}
    \centering
    \includegraphics[width=0.95\linewidth]{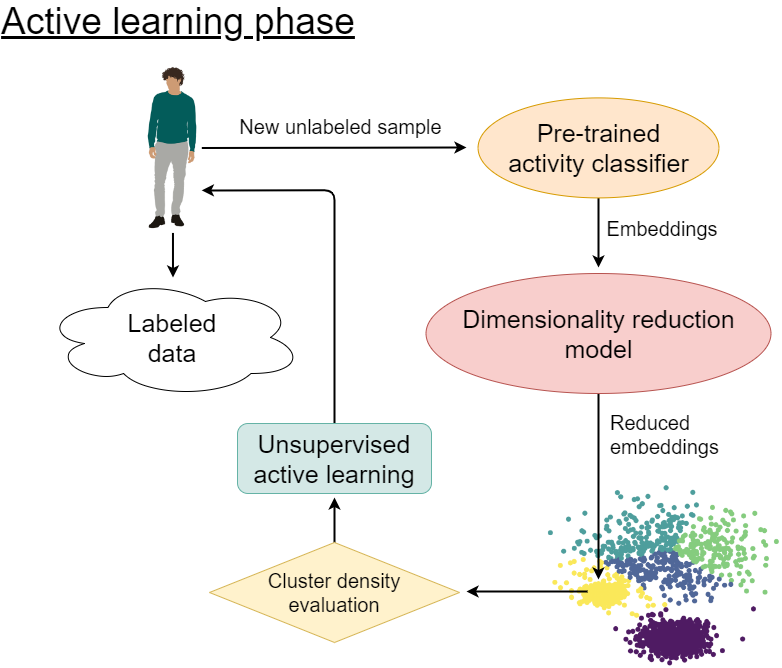}
    \caption{The \textit{active learning phase} of \acronym}
    \label{fig:active_learning_phase}
\end{figure}

As shown in Figure \ref{fig:active_learning_phase},
in the \textit{active learning phase}, each new unlabeled sample collected by the user's mobile device is provided to the pre-trained model, and the result is dimensionally reduced through the personalized dimensionality reduction model. Each new reduced embedding is then mapped to the closest cluster among the ones derived during the \textit{accumulation phase} (e.g., the cluster with the closest centroid). The active learning strategy of \acronym\ asks for a label only for those samples that can be considered as highly representative of the closest cluster (i.e., of each activity pattern) in order to minimize the number of queries that are prompted to the user.


More specifically, an active learning process is started by evaluating whether a new reduced embeddings sample $s$ improves the density of its cluster or not. \acronym\ assigns to each cluster $c$ a threshold $T_c$, which reflects the density level of the cluster. This threshold is automatically computed as the average distance between each pair of points within the cluster.

To determine whether a new sample $s$ is highly representative of cluster $c$, we evaluate the average distance between each pair of points in $c$ after adding $s$ to the cluster. If this average distance is lower than $T_c$, then $s$ is considered highly representative of $c$ and an active learning query is requested to the user. Intuitively, this happens when $s$ is a sample that increases the density of $c$.

Algorithm~\ref{alg:active_learning} shows the pseudo-code of the strategy adopted by \acronym\ to decide if an active learning query is needed.

\begin{algorithm}[]
\footnotesize
\caption{ActiveLearningNeeded}
\label{alg:active_learning}
\hspace*{\algorithmicindent} \textbf{Input:} a new sample $emb'_s$, and its closest cluster $matchingClust$\\
\hspace*{\algorithmicindent} \textbf{Output:} a boolean value 
\begin{algorithmic}[1]

\State $avg \gets $ average distance between points in $matchingClust$
\State $avg' \gets $ average distance between points in $matchingClust \bigcup \{emb_s'\}$

\If{$avg'<avg$}
    \State \Return True
\EndIf
    \State \Return False

\end{algorithmic}
\end{algorithm}

Independently if the active learning query is triggered or not, $s$ is added to $c$,
and $T_c$ is updated accordingly.
The active learning strategy of \acronym\ assumes that the clusters found during the \textit{accumulation phase} are reliable. 
For this reason, as we will describe in Section~\ref{sec:experiments}, we empirically determined $ACC\_TH$, by balancing the trade-off between recognition rate and a number of triggered questions.


\subsection{Fine-tuning phase}
\label{sec:fine_tuning_phase}
Finally, during the \textit{fine-tuning phase}, \acronym\ relies on the data labeled through active learning at the previous phase to fine-tune the deep-learning model pre-trained during the \textit{self-supervised phase}. In this way, \acronym\ builds an activity classifier personalized for each new user of the framework. Specifically, \acronym\ dynamically adds to the \textit{Feature Extraction Layers} of the deep-learning model pre-trained during the \textit{self-supervised phase} a \textit{Fine-Tuning Head} that includes classification layers.
Since it is not possible to know in advance how many activities (and which) the user actually performed, the \textit{Fine-Tuning Head} is created dynamically based on the number of activity classes in the labeled data. During the fine-tuning process, the layers of both the \textit{Feature Extraction Module} and the \textit{Fine-Tuning Head} are unfrozen.

The \textit{fine-tuning phase} should be triggered by considering specific conditions. For instance, it can be a periodic task (e.g., with a time periodicity or each time a certain amount of labels are collected). However, defining when fine-tuning should be started is out of the scope of this paper and it will be investigated in future work.


\section{Experimental evaluation}
\label{sec:experiments}

\begin{table*}[ht!]
\small
\centering
\caption{Main results of \acronym. The \textit{accumulation phase} was performed on 75\% of the test set. The table reports the average number of active learning queries prompted to the user by \acronym\ (i.e., number of samples for fine-tuning), and the average number of labeled samples in the training set used to train \textit{fully supervised}}
\label{tab:main_results}
\begin{tabular}{c|cc|cc}
\hline
Dataset     & \begin{tabular}[c]{@{}c@{}}Average number of\\ active learning queries\\ (fine-tuning samples)\end{tabular} & \acronym\ F1 & \begin{tabular}[c]{@{}c@{}}Average supervised\\ training set size\end{tabular} & \begin{tabular}[c]{@{}c@{}}\textit{Fully supervised}\\ F1\end{tabular} \\ \hline
HHAR        & 369                                                                                            & 0.9734                                   & 25053                                                                          & 0.8792                                                        \\
MotionSense & 143                                                                                            & 0.9291                                   & 5304                                                                           & 0.9483                                                        \\ \hline
\end{tabular}
\end{table*}

This section describes the experiments we carried out to  evaluate \acronym. We first introduce the two publicly available datasets we considered for our experiments. Then, we discuss details about our experimental setup: the data pre-processing, the specific models we used to implement \acronym, and the evaluation methodology we adopted. Finally, we present the results of our evaluation.

\subsection{Datasets}
\subsubsection{HHAR}
The Heterogeneity Human Activity Recognition (HHAR) dataset~\cite{stisen2015smart} includes accelerometer and gyroscope data from $9$ users wearing a smartphone (in a pouch carried on the waist) and a smartwatch. The dataset was collected considering $6$ different activities: \textit{biking}, \textit{sitting}, \textit{standing}, \textit{walking}, \textit{stairs up}, and \textit{stairs down}. HHAR includes heterogeneous sampling rates, based on the specific devices worn by the user. In particular, the dataset includes $2$ different smartwatch devices: LG Watch and Samsung Galaxy Gears with maximum sampling rates of $200Hz$ and $100Hz$, respectively. Moreover, HHAR contains $4$ different smartphone models: Samsung Galaxy S3 mini (maximum sampling rate of $100Hz$), Samsung Galaxy S3 ($150Hz$), LG Nexus 4 ($200Hz$), and Samsung Galaxy S+ ($50Hz$).

\subsubsection{MotionSense}
MotionSense~\cite{malekzadeh2018protecting} is a publicly available HAR dataset collected from $24$ subjects carrying an iPhone 6s in their trousers' front pocket, and performing $6$ different activities: \textit{walking downstairs}, \textit{walking upstairs}, \textit{walking}, \textit{jogging}, \textit{sitting}, and \textit{standing}. The dataset includes both accelerometer and gyroscope data sampled at $50Hz$.

\subsection{Experimental setup}
In the following, we describe our experimental setup.

\subsubsection{Machine learning models}
As self-supervised learning model, in our experiments, we used a version of the \textit{SimCLR} technique~\cite{chen2020simple} adapted for HAR in previous works~\cite{tang2020exploring,haresamudram2022assessing}. Consistently with the previous work, the \textit{Feature Extraction Layers} consist of three 1D convolutional layers with $32$, $64$, and $96$ filters and $24$, $16$, and $8$ kernel sizes, respectively, followed by a global maximum pooling layer. Each couple of consecutive layers is separated by a dropout layer with a dropout rate of $0.1$. 
The \textit{Self-Supervised Head} consists of three fully connected layers with $256$, $128$, and $50$ neurons, respectively. 
The self-supervised model is pre-trained for $50$ epochs and a batch size of $512$ with the SGD optimizer, considering a cosine decay for the learning rate. Moreover, we empirically determined the optimal transformations to apply to the unlabeled data during SimCLR pre-training, finding that, consistently with~\cite{tang2020exploring}, the  application of the rotation transformation only was the best option. 

The \textit{Fine-Tuning Head} consists of a fully connected layer containing $1024$ neurons and a softmax layer for the final classification. Fine-tuning is performed by running $50$ training epochs with a batch size of $1$, using an early stopping technique that stopped the training process when validation loss did not improve for $5$ consecutive epochs, and adopting the Adam optimizer.

As a dimensionality reduction model, we adopted the Uniform Manifold Approximation and Projection (UMAP) technique~\cite{mcinnes2018umap} since previous work demonstrated how its use is effective in considerably improving the performance of clustering algorithms \cite{allaoui2020considerably}. Moreover, since \acronym\ considers a real-time scenario for active learning, it is crucial to train a dimensionality reduction model only at the end of the accumulation phase, and then to map new data samples in the reduced space. These real-time constraints can be addressed by using UMAP, while t-SNE~\cite{van2008visualizing}, for example, would not be appropriate. During our evaluation, we experimentally determined the best hyperparameters for UMAP: $15$ as the number of neighbors, and an euclidean minimum distance equal to $0.4$.

Realistic and real-time constraints mainly guided the choice of the clustering algorithm of \acronym. First of all, it is not realistic to assume knowing the exact number of different activity patterns that the user will perform while using \acronym. Thus, it would not be 
appropriate to consider clustering algorithms  that require knowing the number of clusters in advance (e.g., K-means). Moreover, in real-time scenarios, we need to rely on an algorithm that (i) clusters the reduced embeddings accumulated during the \textit{accumulation phase}, and then (ii) directly maps new data into the existing clusters during the \textit{active learning phase}. For these reasons, in our experiments, we relied on the HDBSCAN clustering algorithm~\cite{mcinnes2017hdbscan}. In particular, we empirically determined its hyperparameters: a minimum cluster size of $6$, and a minimum number of samples of $10$.

\subsubsection{Pre-processing}
Since the version of SimCLR we are using only considers data from a single tri-axial sensor as input, we focus on the accelerometer data collected through the users' smartphones. Such data are segmented into fixed-length windows with a $50\%$ of overlap. In particular, following the evaluation protocol proposed in~\cite{tang2020exploring}, we considered windows of $400$ samples for MotionSense. On the other hand, since HHAR includes data collected with different sampling rates, we experimentally determined the size of the windows, obtaining a window size of $800$ samples. Due to unsatisfying recognition rates during preliminary experiments, in MotionSense we grouped the labels \textit{walking downstairs} and \textit{walking upstairs} into the label \textit{walking stairs}, while in HHAR we grouped \textit{stairs down} and \textit{stairs up} into \textit{stairs}. Indeed, these activities are particularly challenging to discriminate.

\subsubsection{Baseline}
In our evaluation, we compared \acronym\ with a \textit{fully supervised} approach: a deep neural network with the same architecture of the network used by \acronym\ during the \textit{fine-tuning phase}, trained in a supervised way considering full availability of labeled training data.

\subsubsection{Evaluation Methodology}
\label{sec:evametho}
The \acronym\ framework is designed for personalizing the HAR model for each user by leveraging a limited amount of labeled data.
Hence, we performed our experiments considering the leave-one-subject-out cross-validation technique. At each iteration of the cross-validation process, data from one user is used as the test set (i.e., accumulation, active learning, and fine-tuning phases), while the remaining folds are used as the training set (i,e., self-supervised phase). At the end of the process, we averaged the results obtained at each iteration in terms of weighted F1 score.

Considering the \textit{fully supervised} baseline, at each fold, $10\%$ of the training set was used as the validation set. 
On the other hand, in order to simulate the \textit{accumulation} and \textit{active learning} phases of \acronym,
a certain percentage (i.e. \textit{accumulation threshold}) of the test set was used to simulate the samples that are accumulated by \acronym\ during the \textit{accumulation phase}. The remaining samples of the test set were used to simulate the data stream processed during the \textit{active learning phase}\footnote{This setting resembles a scenario in which practitioners must allocate both \textit{accumulation} and the \textit{active learning} phases within a given timeframe (e.g., one month) that is constrained by deployment or project requirements.}.

During fine-tuning, we used $10\%$ of labeled data as the validation set. 
Finally, all the samples in the test set that were not labeled (i.e., the ones collected in the \textit{accumulation phase} and the samples not labeled during the \textit{active learning phase}) were used to assess the recognition rate of the fine-tuned model. We also compute the \textit{active learning rate}, i.e., the ratio between the amount of triggered active learning questions and the total number of samples available during the \textit{active learning phase}.

While it was straightforward to apply the leave-one-out cross-validation on HHAR, it was more challenging on MotionSense. Indeed, users in this dataset do not have enough data samples to obtain meaningful results from the \textit{accumulation} and \textit{active learning} phases. Hence, we created groups of $4$ or $5$ users performing activities in a similar way (based on available meta-data like age, height, weight, etcetera) and we considered each group as a single user. This resulted in $5$ pseudo-users that we used to perform the leave-one-out cross-validation.


\begin{figure*}[ht]
     \centering
     \begin{subfigure}[t]{0.3\linewidth}
         \centering
         \includegraphics[width=0.95\linewidth]{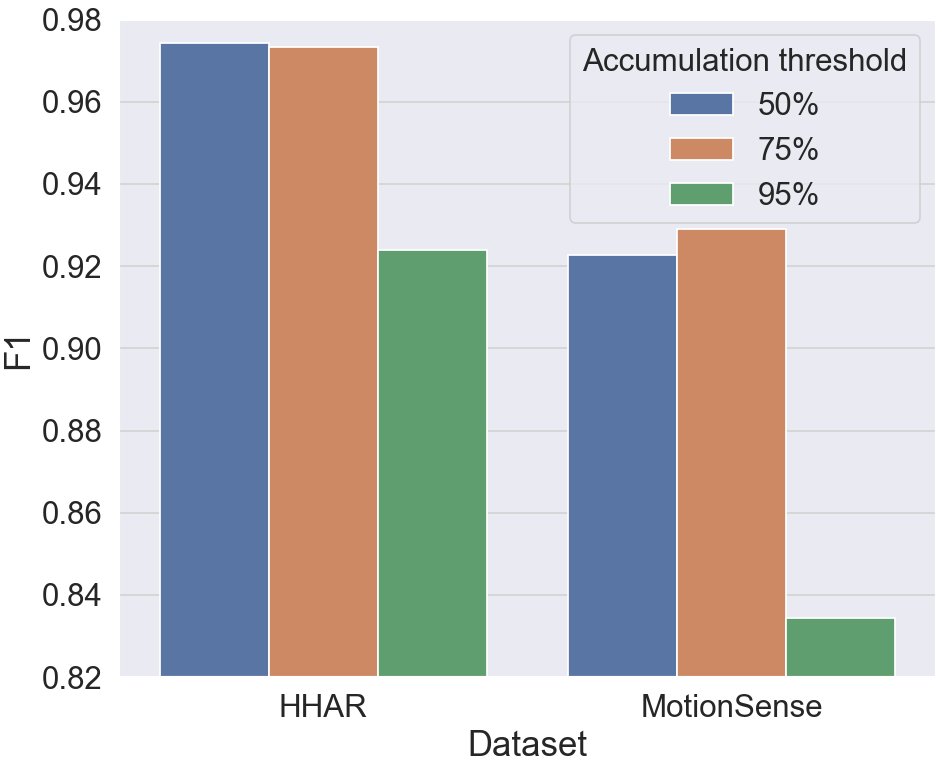}
        \caption{Recognition rates of \acronym\ by varying the \textit{accumulation threshold}\newline}
        \label{fig:f1_results}
     \end{subfigure}
     \hfill
     \begin{subfigure}[t]{0.3\linewidth}
         \centering
         \includegraphics[width=0.97\linewidth]{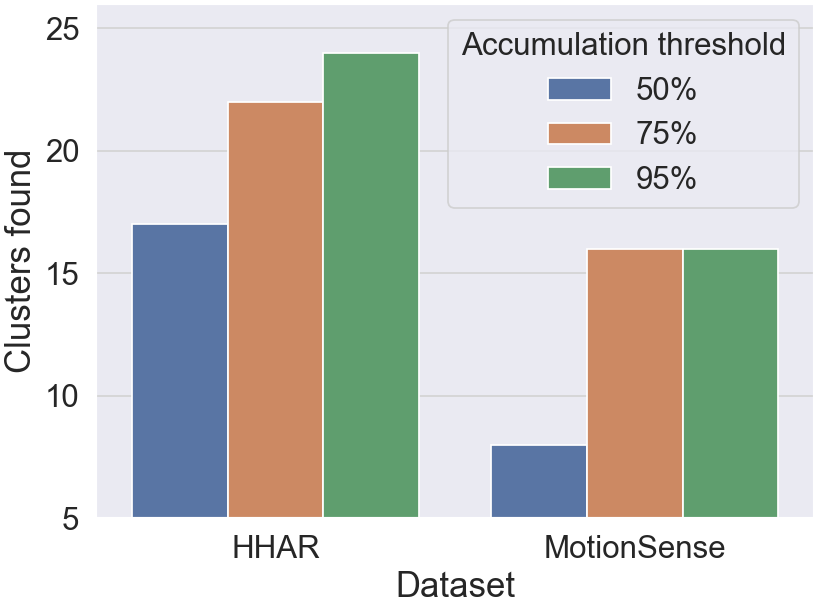}
        \caption{Number of clusters derived by \acronym\ at the end of the \textit{accumulation phase} by varying the \textit{accumulation threshold}}
        \label{fig:clusters_found_results}
     \end{subfigure}
     \hfill
     \begin{subfigure}[t]{0.3\linewidth}
         \centering
         \includegraphics[width=\linewidth]{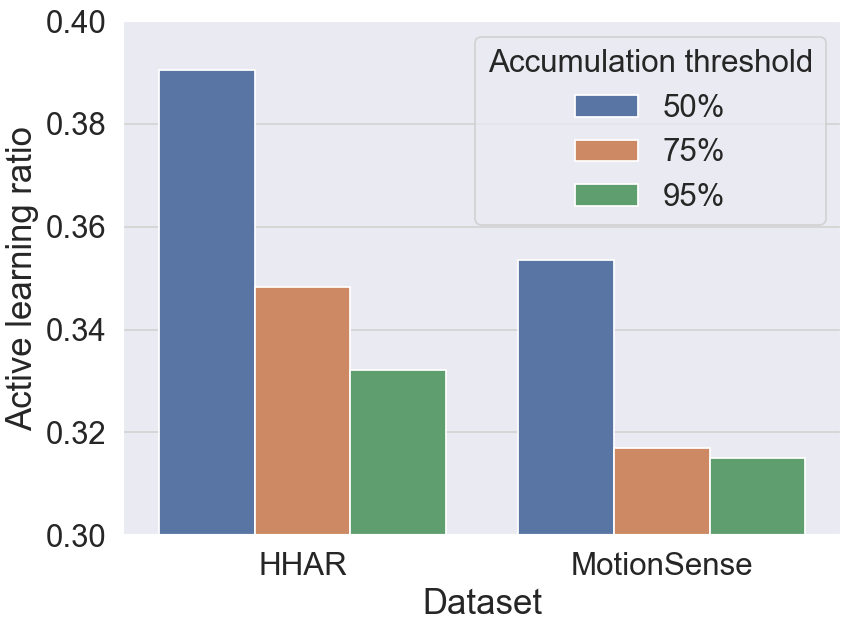}
        \caption{Active learning rate during the \textit{active learning phase} of \acronym\ by varying \textit{accumulation threshold}}
        \label{fig:active_learning_frequency_results}
     \end{subfigure} 
        
        \caption{The impact of \textit{accumulation threshold} on recognition rate, number of clusters, and active learning rate}
        \label{fig:accumulation}
\end{figure*}

\subsection{Results}
In the following, we describe the results we obtained during our experimental evaluation. 

\subsubsection{Recognition rate}
Table \ref{tab:main_results} compares \acronym\ and the \textit{fully supervised} baseline in terms of F1 score and the average number of necessary labeled data to deploy such solutions on both datasets. Such results were obtained by accumulating during the \textit{accumulation phase} a number of samples that corresponds to $75\%$ of the data available in the test set. Considering HHAR, \acronym\ dramatically outperforms \textit{fully supervised} in terms of F1 score (i.e., $\approx +10\%$). Moreover, this result is obtained considering only $369$ labeled samples per user, while the \textit{fully supervised} baseline, at each iteration of the cross-validation, relied on average on $\approx 25k$ labeled samples.

On the other hand, considering MotionSense, our framework did not improve the recognition rates of \textit{fully supervised}. This happens since the \textit{active learning phase} of \acronym\ is actually computed on the data of more users (see Section~\ref{sec:evametho}). Hence, with a few fine-tuning samples labeled with active learning, it is hard to cover all the possible ways in which an activity can be performed by different users. Nonetheless, the F1 score reached by \acronym\ is only $\approx 2\%$ behind \textit{fully supervised}, but relying only on $143$ labeled samples per group of users included in the test set, instead of considering on average more than $5k$ labeled training samples. Hence, we believe that these results still highlight the effectiveness of \acronym, which 
reaches recognition rates close to supervised approaches with a limited amount of labeled samples.


\subsubsection{Impact of accumulation threshold}
Figure~\ref{fig:accumulation} depicts how the \textit{accumulation threshold} affects the performance of \acronym. 
By setting the \textit{accumulation threshold} at $50\%$ (i.e., $50\%$ of data used for accumulation and $50\%$ of data in the active learning phase) we get a high recognition rate, at the cost of having the highest active learning rate. 
Indeed, with this threshold, \acronym\ finds a lower number of sparse clusters, thus requiring a higher number of samples during the active learning phase to increase their densities. This also implies a higher number of active learning queries.

On the other hand, when the accumulation threshold is $95\%$ (i.e., $95\%$ of data used for accumulation and $5\%$ of data in the active learning phase) the recognition rate is the lowest. This is 
due to the fact that this setting does not 
provide a sufficient amount of data in the active learning phase for properly fine-tuning the model.

Hence, we observed that setting the threshold to $75\%$  (i.e., $75\%$ of data used for accumulation and $25\%$ of data in the active learning phase) was the best trade-off between the recognition rate (that is similar to the one obtained with the threshold set to $50\%$) with a significantly reduced active learning rate. Indeed, this setting allows \acronym\ to identify clusters that better reflect activity patterns and that can be effectively used for the active learning phase. 

In general, our results show that increasing the accumulation threshold decreases the active learning rate. This is due to the fact that our unsupervised learning strategy is based on cluster density. Indeed, the higher the amount of accumulated data and the denser the clusters.

\section{Conclusion and Future Work}
In this paper, we presented \acronym: a framework for personalized sensor-based HAR. Thanks to a novel combination of self-supervised learning and active learning, \acronym\ requires only a limited amount of labeled data from target users to fine-tune a model pre-trained with unlabeled data only.
While the results obtained so far are promising, we plan to investigate several aspects in future work.

First, even though \acronym\ does not assume knowing in advance the activities performed by the target user, in this work the pre-trained feature extractor and fine-tuning are trained considering the same activity classes. In future work, we want to truly assess the capability of \acronym\ for activity discovery and domain adaptation. A challenge will be to distinguish outliers from new activities.

A drawback of \acronym\ is that it relies on a fixed threshold to decide when switching from \textit{accumulation phase} to \textit{active learning phase}. In future work, we will investigate automatic methods based on the silhouette score of the clusters.

Another limitation of \acronym\ is that, while it requires limited labeled data samples to obtain high recognition rates, the active learning query rate is relatively high. This aspect has a negative impact on usability. We will investigate more sophisticated active learning strategies aiming at further minimizing the number of required samples. For instance, an interesting strategy is to establish a budget of queries for each cluster.
Nonetheless, these results are also probably due to the fact that the considered datasets are small, and we could not evaluate \acronym\ with large amounts of data for the accumulation and active learning phases. We expect that, in real-world scenarios where it is possible to collect a large amount of unlabeled data, the active learning rate would be significantly lower.

\acronym\ also assumes that it is possible to store and process locally unlabeled and labeled datasets. However, in real-world scenarios, unlabeled data may be collected for a significantly long time. For instance, the accumulation phase may last several days.
We will investigate solutions based on trusted edge devices with storage and computational capabilities. Users' devices should be only in charge of running the fine-tuned classifier.

\section*{Acknowledgment}

This work was partially supported by the project ``MUSA - Multilayered Urban Sustainability Action'',  NextGeneration EU, MUR PNRR.

\bibliographystyle{ieeetr}
\bibliography{references}

\end{document}